\documentclass[11pt,a4paper]{article}
\usepackage[hyperref]{emnlp-ijcnlp-2019}
\usepackage{times}
\usepackage{latexsym}
\usepackage{tabularx}
\usepackage{amsfonts}
\usepackage{booktabs}
\usepackage{afterpage}
\usepackage{microtype}
\usepackage{xcolor}
\usepackage{graphicx}
\usepackage{subcaption}
\usepackage{paralist}
\usepackage{url,bm}
\usepackage{amsmath, cancel}
\usepackage{multirow}
\usepackage{url}
\aclfinalcopy 

\title{Examining Gender Bias in Languages with Grammatical Gender}

\author{Pei Zhou$^{1,2}$, Weijia Shi$^1$, Jieyu Zhao$^1$, Kuan-Hao Huang$^1$,\\ \textbf{Muhao Chen}$^{1,3}$, \textbf{Ryan Cotterell}$^4$, \textbf{Kai-Wei Chang}$^1$\\
  $^1$Department of Computer Science, University of California Los Angeles \\
  $^2$Department of Computer Science, University of Southern California \\
  $^3$Department of Computer and Information Science, University of Pennsylvania \\
  $^4$Department of Computer Science, Johns Hopkins University \\
  {\tt peiz@usc.edu};
  {\tt \{swj0419,jyzhao, khhuang, kwchang\}@cs.ucla.edu}; \\
  {\tt muhao@seas.upenn.edu};
  {\tt ryan.cotterell@jhu.edu}
}

\date{}

\begin{document}
\maketitle
\begin{abstract}
Recent studies have shown that word embeddings exhibit gender bias inherited from the training corpora. However, most studies to date have focused on quantifying and mitigating such bias only in English. These analyses cannot be directly extended to languages that exhibit morphological agreement on gender, such as Spanish and French. In this paper, we propose new metrics for evaluating gender bias in word embeddings of these languages and further demonstrate evidence of gender bias in bilingual embeddings which align these languages with English. Finally, we extend an existing approach to mitigate gender bias in word embeddings under both monolingual and bilingual settings.
Experiments on modified Word Embedding Association Test, word similarity, word translation, and word pair translation tasks show that the proposed approaches effectively reduce the gender bias while preserving the utility of the embeddings.

\end{abstract}

\section{Introduction}\label{intro}
Word embeddings are widely used in modern natural language processing tools. By virtue of their being trained on large, human-written corpora, recent studies have shown that word embeddings, in addition to capturing
a word's semantics, also encode gender bias in society~\cite{kaiwei,caliskan2017semantics}. As a result of this bias, the embeddings may cause undesired consequences in the resulting models~\cite{ZWYOC18, font2019equalizing}. Therefore, extensive effort has been put toward analyzing and mitigating gender bias in word embeddings \cite{kaiwei,zhao2018learning,dev2019attenuating,ethayarajh2019understanding}.

Existing studies on gender bias almost exclusively focus on English (EN) word embeddings.
Unfortunately, the techniques used to mitigate bias in English word embeddings cannot be directly applied to languages with
grammatical gender\footnote{Grammatical gender is a complicated linguistic phenomenon; many languages contain more than two gender classes. In this paper, we focus on languages with masculine and feminine classes. For gender in semantics, we follow the literature and address only binary gender.}, where all nouns are assigned a gender class and the corresponding dependent articles, adjectives, and verbs must agree in gender with the noun (e.g. in Spanish: \textit{la buena enfermera} the good female nurse, \textit{el buen enfermero} the good male nurse)~\cite{corbett1991gender,corbett2006agreement}.
This is because most existing approaches define bias in word embeddings based on the projection of a word on a gender direction (e.g. ``nurse'' in English is biased because its projection on the gender direction inclines towards  female). When the grammatical gender exists, such bias definition is problematic as masculine and feminine words naturally carry gender information. For example, ``enfermero'' (male nurse) and ``enfermera'' (female nurse) lean toward male and female, respectively. This is due to morphological agreement and should not be considered as a stereotype. 

However, gender bias in the embeddings of languages with grammatical gender indeed exists. For example, when we align Spanish (ES) embeddings to English embeddings, the word ``abogado'' (male lawyer)  is closer to ``lawyer'' than ``abogada'' (female lawyer).
This observation implies a discrepancy in semantics between the masculine and feminine forms of the same occupation in Spanish word embeddings. A similar observation is also found in other languages, such as French (FR). 


In this paper, we refer languages that have gender distinctions for all nouns (e.g., Spanish and German) as \emph{gendered languages} and others that do not mark grammatical gender as \emph{languages without grammatical gender} (e.g., English). 
We use Spanish and French as running examples and propose quantitative methods for evaluating bias in word embeddings of gendered languages and bilingual word embeddings that align a gendered language with English. 
We first define gender bias in the word embeddings of gendered languages by constructing two gender directions: the semantic gender direction and the grammatical gender direction. We then analyze gender bias in bilingual embeddings using a similar approach.



To mitigate gender bias in these embeddings, we propose two approaches. One is shifting words along the semantic gender direction with respect to an anchor point and the other is mitigating English first and then aligning the embedding spaces. Results show that a hybrid of these two approaches effectively mitigate bias in Spanish and French word embeddings as well as ES-EN, FR-EN bilingual word embeddings. We also show through word similarity and word translation experiments that the utility of the original monolingual and bilingual word embeddings is preserved.

Our contributions are summarized in the following. (1) We show that word embeddings of gendered languages such as Spanish and French contain gender bias and bilingual word embeddings aligning these languages to English also inherit the bias. (2) Based on the observation, we propose new definitions of gender bias by constructing two gender directions. (3) We propose methods to reduce gender bias for both monolingual and bilingual embeddings and show that they effectively mitigate bias while preserving the original utility of embeddings. Source code and data are available at  \url{https://github.com/shaoxia57/Bias_in_Gendered_Languages}.

\section{Gender Bias Analysis}\label{analysis}
\label{sec:mweat}
This section discusses how to analyze bias in embeddings of gendered languages and bilingual word embeddings.  

\subsection{Bias in Gendered Languages} \label{bias_ggl}
In gendered languages, all nouns are assigned a gender class. However, inanimate objects (e.g., water and spoon) do not carry the meaning of male or female. 
To address this issue, we define two gender directions: (1) \textit{semantic gender}, which is defined by a set of gender definition words (e.g., man, male, waitress) and (2) \textit{grammatical gender}, which is defined by a set of masculine and feminine nouns (e.g., water, table, woman). Most analysis of gender bias assumes that the language (e.g,. English) only has the former direction. However, when analyzing languages like Spanish and French, considering the second is necessary. We discuss these two directions in detail as follows.

%

\paragraph{Semantic Gender} Following~\citet{kaiwei}, we collect a set of gender-definition pairs (e.g., ``mujer'' (woman) and ``hombre'' (man) in Spanish). Then, the gender direction is derived by conducting principal component analysis (PCA)~\cite{jolliffe2011principal} over the differences between male- and female-definition word vectors. Similar to the analysis in English, we observe that there is one major principal component carrying the meaning of gender in French and Spanish and we define it as the semantic gender direction $\vec{d}_{PCA}$.

\paragraph{Grammatical Gender} 
The number of grammatical gender classes ranges from two to several tens~\cite{corbett1991gender}. To simplify the discussion, we focus on noun class systems where two major gender classes are feminine and masculine. However, the proposed approach can be generalized to languages with multiple gender classes (e.g., German). To identify grammatical gender direction, we collect around 3,000 common nouns that are grammatically masculine and 3,000 nouns that are feminine.
Since most nouns (e.g., water) do not have a paired word in the other gender class, we do not have pairs of words to represent different grammatical genders. Therefore, instead of applying PCA, we learn the grammatical gender direction $\vec{d_g}$ by Linear Discriminant Analysis (LDA)~\cite{fisher1936use}, which is a standard approach for supervised dimension reduction. The model achieves an average accuracy of 0.92 for predicting the grammatical gender in Spanish and 0.83 in French with 5-fold cross-validation. 


Comparing $\vec{d}_{PCA}$ with $\vec{d}_g$, the cosine similarity between them is 0.389, indicating these two directions are overlapped to some extent but not identical. This is reasonable because words such as ``mujer (woman)'', ``doctor (female doctor)'' are both semantically and grammatically marked as feminine. 
To better distinguish between these two directions, we project out the grammatical gender component in the computed gender direction to make the semantic gender direction $\vec{d}_s$ orthogonal to the grammatical gender direction:
\begin{equation*}
    \vec{d_s} = \vec{d}_{PCA} - \left \langle \vec{d}_{PCA} , \vec{d}_{g} \right \rangle \vec{d}_{g},
\end{equation*}
where $\left \langle \vec{x},\vec{y}  \right \rangle$ represents the inner product of two vectors. 
\paragraph{Visualizing and Analyzing Bias in Spanish}
We take Spanish as an example and analyze gender bias in Spanish fastText word embeddings~\cite{fasttext} pre-trained on Spanish Wikipedia. For simplicity, we assume that the embeddings contain all gender forms of the words.
To show bias, we randomly select several pairs of gender-definition and occupation words as well as inanimate nouns in Spanish and visualize them on the two gender directions defined above.

\begin{figure}
	\includegraphics[width=\columnwidth]{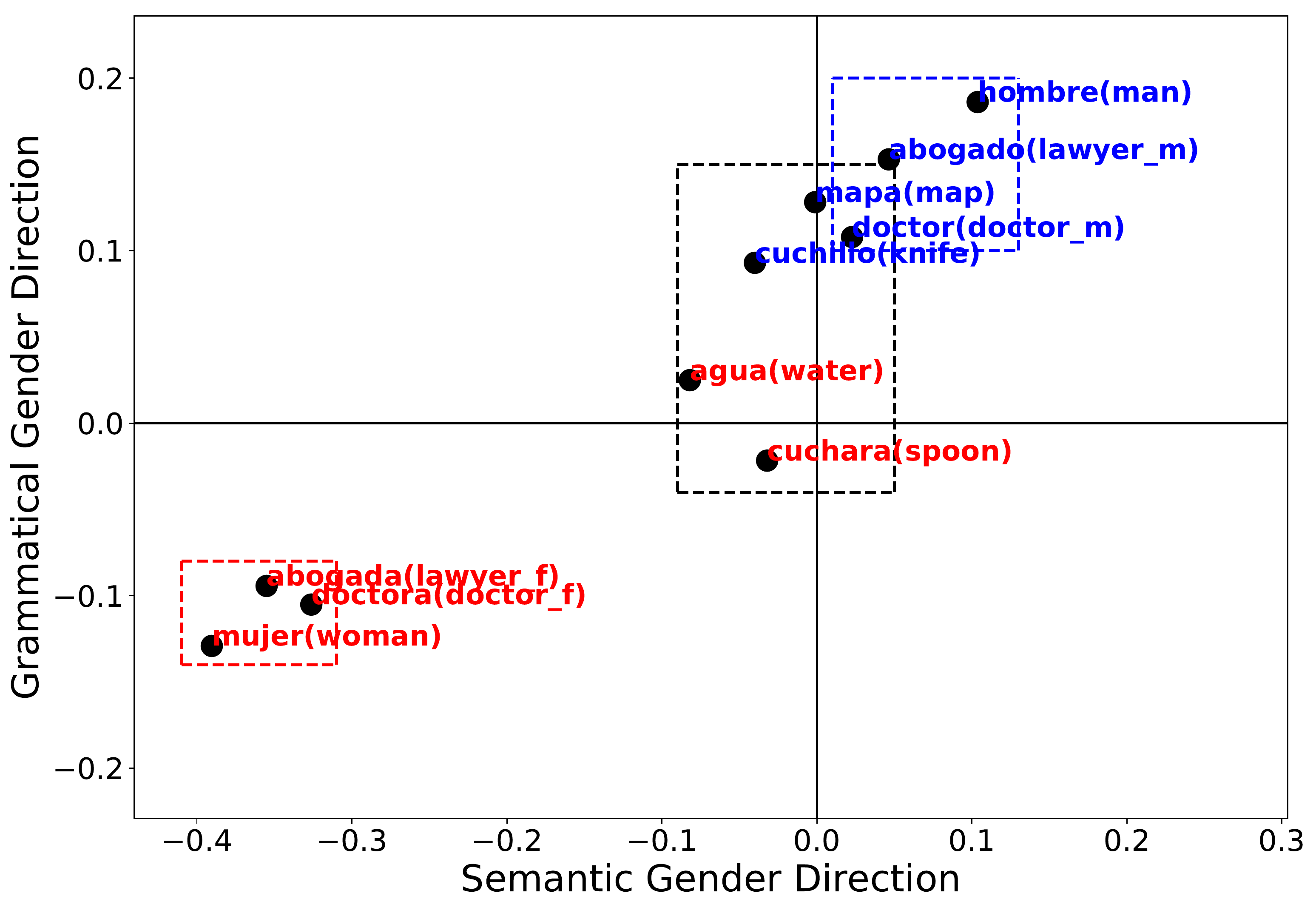}
	\centering
	\caption{Projections of selected words in Spanish on grammatical and semantic directions with masculine nouns in blue and feminine nouns in red. We find that: (1) Most inanimate nouns (e.g., map, spoon; enclosed by black dotted lines) lie near the origin point of the semantic gender axis; (2) Masculine definition and occupation words (blue dotted lines) lie in the masculine side for both gender directions and so do feminine words (red dotted lines). But the feminine words are farther on the feminine side on the semantic gender direction compared to masculine words.}~\label{fig:viz_2directions}
\end{figure}

Figure \ref{fig:viz_2directions} shows that the inanimate nouns lie near the origin point on the semantic gender direction, while the masculine and feminine forms of the occupation words are on the opposite sides for both directions.
However, while projections of occupation words on the grammatical gender direction are symmetric with respect to the origin point, they are asymmetric when projected on the semantic gender direction. 
Along the semantic gender direction, female occupation words incline more to the feminine side than the male occupation words to the masculine side. This discrepancy shows that the embeddings carry unequal information for two genders. 



\paragraph{Quantification of Gender Bias}
To quantify gender bias in English embeddings, \citeauthor{caliskan2017semantics}~(\citeyear{caliskan2017semantics}) propose Word Embedding Association Test (WEAT), which measures the association between two sets of target concepts and two sets of attributes. Let $X$ and $Y$ be equal-sized sets of target concept embeddings (e.g., words related to mathematics and art) and let $A$ and $B$ be sets of attribute embeddings (e.g., male and female definition words such as ``man'', ``girl''). Let $\cos(\vec{a},\vec{b})$ denote the cosine similarity between vectors $\vec{a}$ and $\vec{b}$. The test statistic is a difference between sums over the respective target concepts,  \begin{equation*}
    s(X,Y,A,B)=\sum_{\vec{x}\in X}s(\vec{x},A,B) - \sum_{\vec{y}\in Y}s(\vec{y},A,B),
\end{equation*}
where $s(\vec{w},A,B)$ is the difference between average cosine similarities of the respective attributes,
 \begin{equation*}
         s(\vec{w},A,B) \!=\! \frac{1}{|A|}\!\sum_{\vec{a} \in A} \cos (\vec{w},\vec{a}) - \frac{1}{|B|}\!\sum_{\vec{b} \in B} \cos (\vec{w},\vec{b}).
 \end{equation*}

In the following, we extend the definition to quantify individual words in gendered languages. 
We consider two types of words. One is \textit{inanimate nouns}, like ``agua'' (water, feminine). They are assigned as either a masculine or feminine noun.  The other is \textit{animate nouns} that usually have two gender forms, like ``doctor'' (male doctor) and ``doctora'' (female doctor).\footnote{Note that some animate nouns in certain languages, like ``m\'{e}decin'' (male doctor) in French, only have one gender form. We omit such cases in this paper for the sake of simplicity. }  

For inanimate nouns that should not be semantically leaning towards one gender, we can quantify the gender bias for word $w$ simply using WEAT,
\begin{equation*}
    \text{b}_w = \left |s(\vec{w},A,B)\right|,
\end{equation*}
which measures the association strength of the word $w$ with the gender concepts. 
The larger $\text{b}_w$ is, the stronger its association with the gender concepts.

For nouns with two gender forms, we test if their masculine ($\vec{w}_m$) and feminine ($\vec{w}_f$) forms are symmetrical with respect to gender definition terms.
\begin{equation}
\label{eq:mweat}
    \text{b}_w = \left |\left |s(\vec{w}_m,A,B)\right | - \left |s(\vec{w}_f,A,B)\right |\right |.
\end{equation}
The larger the value is, the larger gender bias presents. 
For example, if ``doctor'' (male doctor) leans more toward to masculine (i.e., close to masculine definition words like ``el'' (he) and far away from feminine definition words like ``ella'' (she)) than ``doctora'' leans toward to feminine, then embeddings of ``doctor'' and ``doctora'' are biased. We will discuss detailed quantitative analysis using this metric in Section~\ref{experiments}.

\begin{figure}
	\includegraphics[width=\columnwidth]{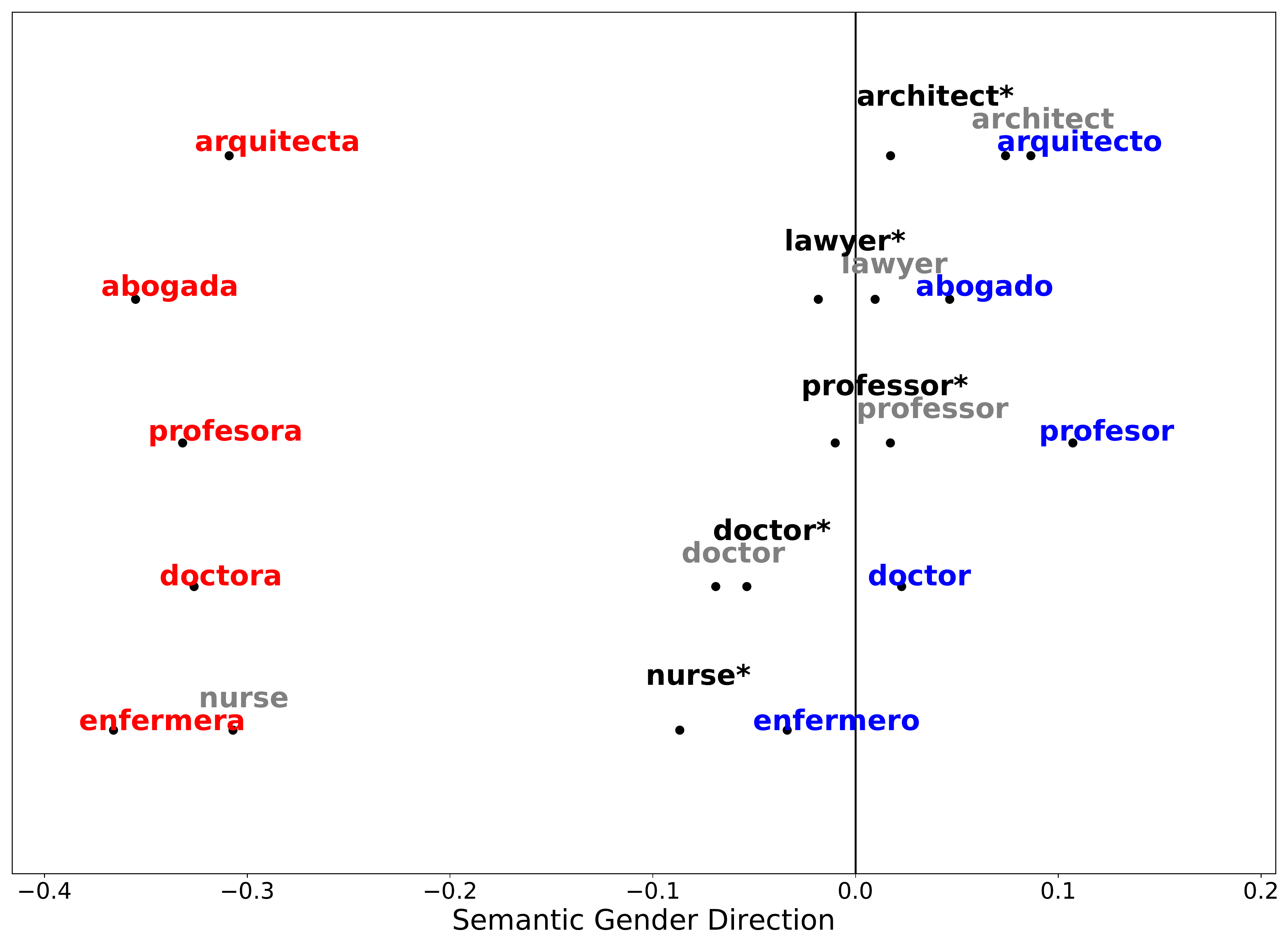}
	\centering
	\caption{Projections of occupation words on the semantic gender direction in two gender forms (blue for male and red for female) in Spanish, the original English word (grey), and the bias-mitigated version of the English word (black with ``*''). We find that feminine occupations are farther away from the English counterparts as well as the neutral position (indicated by the vertical line) compared to masculine words.}~\label{fig:viz_bilingual_before}
\end{figure}
\subsection{Bias in Bilingual Word Embeddings}
Bilingual word embeddings are used widely in tasks such as multilingual transfer. It is essential to quantify and mitigate gender bias in these embeddings to avoid them from affecting downstream applications. We focus on bilingual word embeddings that align a grammatical gender language like Spanish with a language without grammatical gender like English. The definitions of the two gender directions are similar: we construct the grammatical gender direction using the same sets of words in the gendered language since the other language does not mark grammatical gender. We combine the gender-definition words for both languages and remove the grammatical gender component to get the semantic gender direction similarly. 

We use bilingual word embeddings from MUSE~\cite{conneau2017word} that align English and Spanish fastText embeddings together in a single vector space. Figure~\ref{fig:viz_bilingual_before} shows the projections of 5 occupation words on the semantic gender direction. First, we notice that the embeddings of some English occupations are much closer to one gender form than the other in Spanish. For example, ``nurse'' in English is closer to ``enfermera'' (female nurse) than ``enfermero'' (male nurse) in Spanish. We further separate out the bias in English embedding in the analysis by mitigating the English embeddings using ``hard-debiasing'' approach proposed in~\citet{kaiwei} and aligning Spanish embedding with the bias-reduced English embedding (words annotated with ``*''). Results in Figure~\ref{fig:viz_bilingual_before} demonstrate another type of gender unequal in bilingual embedding: 
masculine occupation words are closer to the corresponding English occupation words than the feminine counterparts. This result aligns with the observation in Figure~\ref{fig:viz_2directions}. 



More quantitative analyses about gender bias in bilingual embedding will be discussed in 
Section~\ref{bilingual_exp}.

\section{Mitigation Methods}\label{mitigation}

In the following, we first describe the optimization objectives for mitigating gender bias. Then, we present mitigation methods under two settings: (1) \emph{monolingual setting}, where we only have access to word embeddings of the gendered language, and (2) \emph{bilingual setting}, where the embeddings of a language without grammatical gender (e.g. English) aligned in the same space are also provided.
Specifically, for the bilingual setting, we can use English word vectors to facilitate the bias mitigation.
\subsection{Mitigation Objectives}
For inanimate nouns, we project out their semantic gender component as they should not carry semantic gender. Equivalently, we minimize the inner product 

\begin{equation}
\label{eq:inanimate}
\left \langle \vec{w},\vec{d_s}  \right \rangle
\end{equation} between the inanimate word vector $\vec{w}$ and the semantic gender direction $\vec{d_s}$.

For nouns with both gender forms, we aim at removing the gender inequality on the semantic gender direction as shown in Figure~\ref{fig:viz_2directions} and Figure~\ref{fig:viz_bilingual_before}. Specifically, we minimize the difference between the distance of feminine form of the word $\vec{w_f}$ and the masculine form of the word $\vec{w_m}$ to an anchor point (e.g., the origin point) on the semantic gender direciton $d_s$:
\begin{equation}
\label{eq:objmf}
     \left | \left | \left \langle \vec{w}_m,\vec{d}_s  \right \rangle \!-\! \left \langle \vec{w}_a,\vec{d}_s  \right \rangle \right | \!-\!  \left | \left \langle \vec{w}_f,\vec{d}_s  \right \rangle \!-\! \left \langle \vec{w}_a,\vec{d}_s  \right \rangle \right | \right |.
\end{equation}
Note that for most cases, $\vec{w_m}$ and $\vec{w_f}$ lie on the opposite sides of the anchor point on the semantic gender axis. Therefore, we can simplify Eq. \eqref{eq:objmf} as
\begin{equation}
\label{eq:biasobj}
     \left | \left \langle \vec{w}_m,\vec{d}_s  \right \rangle + \left \langle \vec{w}_f,\vec{d}_s  \right \rangle - 2 \left \langle \vec{w}_a,\vec{d}_s  \right \rangle \right |.
\end{equation}


Intuitively, this measures how much we need to move the word pair on the semantic gender direction so that they are symmetric with respect to the anchor point. If the anchor point is the origin point (i.e. $\left \langle \vec{w}_a,\vec{d}_s  \right \rangle=0$),  this objective is the absolute value of the sum of two projections.
\subsection{Monolingual Setting}
\paragraph{Shifting Along Semantic Gender Direction (Shift)}
We mitigate bias by minimizing Eq. \eqref{eq:biasobj} and Eq. \eqref{eq:inanimate} in a post-processing manner. 
Effectively, for inanimate nouns, we remove the semantic gender component from the embeddings. For nouns with two gender forms (e.g., occupations),  
we shift the two forms along the semantic gender direction so that they have the same distance to the anchor point.
For monolingual setting, we use the origin point on the gender direction as the anchor point and call the method Shift\_Ori.

Note that \citet{gonen2019lipstick} show that mitigating gender bias in word embeddings by moving on the gender direction is not sufficient and words with gender bias still tend to group together. Similairly, our approach might not entirely remove the gender bias. However, the approach in \citet{gonen2019lipstick} to test the remainder bias cannot be applied directly for gendered languages, because grouping of the embeddings of masculine and feminine words does not always indicate bias due to grammatical gender .

\subsection{Bilingual Setting}
\paragraph{Mitigating Before Alignment (De-Align)}
As gender bias can be observed when aligning gendered languages with language without grammatical gender, we consider using English to facilitate mitigating bias. Specifically, we first apply the ``hard-debasing'' approach~\citet{kaiwei} to English embedding and then align gendered language with the bias-reduced English. Because most words with both gender forms align to the same word in English (e.g., ``enfermero'' and ``enfermera'' in Spanish both align to ``nurse'' in English), the alignment places the two gender forms of the words in a more symmetric positions in the vector space after the alignment. 
\paragraph{Shifting Along Semantic Gender Direction (Shift)}
We consider using English words as anchor positions when mitigating bias using Eq \eqref{eq:biasobj}. We call this approach Shift\_EN. For example, when applying Eq. \eqref{eq:biasobj} to re-position ``enfermero'' and ``enfermera'' in Spanish we take English word ``nurse'' aligned in the same space as the anchor position. 

\paragraph{Hybrid Method (Hybrid)}
We also consider a hybrid method that integrates the aforementioned two approaches.
In particular, we first mitigate English embeddings, align English embeddings with the embeddings of gendered languages, and then shift words in languages with grammatical gender along the semantic gender direction. 
We consider two variants based on how the anchor positions are chosen: (1) Hybrid\_Ori uses the origin point as the anchor position; and (2) Hybrid\_EN uses the corresponding bias-reduced English word as an anchor position when shifting a pair of words with different gender forms. 




\section{Experiments}\label{experiments}
\begin{table*}[ht]
\centering
\begin{tabularx}{\linewidth}{c|c||c||c|c|c|c|c}
Lang. & Metric         & Original       & Shift\_Ori    & Shift\_EN     & De-Align & Hybrid\_Ori    & Hybrid\_EN \\ \hline
ES & MWEAT--Diff      & 3.6918          & 0.3090          & 0.3324 & 3.5748    & \textbf{0.3090}          & 2.2494      \\ 
ES & MWEAT--p-value    & 0.0000          & 0.1130          & 0.0010  & 0.0010   & \textbf{0.7330}          & 0.0020     \\ \hline
FR & MWEAT--Diff      & 2.3437          & 0.2446 & 0.3882          & 2.3436    & \textbf{0.2446} & 1.1758      \\ 
FR & MWEAT--p-value    & 0.0000          & 0.1470 & 0.0010          & 0.0020    & \textbf{0.5290} & 0.0910      \\ \hline
ES & Word Similarity & \textbf{0.7392} & 0.7363          & 0.7359         & \textbf{0.7392}    & 0.7358          & 0.7356      \\
FR & Word Similarity & \textbf{0.7294} & 0.7218          & 0.7218         & 0.7156    & 0.7218          & 0.7218     
\end{tabularx}
\caption{Analyses on Spanish and French monolingual embeddings before and after bias mitigation. Results show that the original Spanish and French embeddings exhibit strong bias and Hybrid\_Ori significantly reduces the bias in the embedding to to an insignificant level (p-value $>$ 0.05).}
\label{tab:mono_results}
\end{table*}

We evaluate gender bias mitigation methods in both monolingual and bilingual settings and 
test the utility of the bias-reduced embeddings in word-level translation tasks~\citet{conneau2017word}.
Following other analyses on gender bias~\cite{kaiwei,caliskan2017semantics}, we quantify the gender bias based on a set of animate nouns including occupations and gender-definition words. These words are representative to the major challenge when quantifying and mitigating gender bias.
Different from previous analyses in English, most occupation words in gendered languages have more than one forms and their embeddings are inherently different from each other due to morphological agreement. 


\subsection{Data and Configuration}
We apply the proposed mitigation methods on Spanish-English and French-English bilingual embeddings from MUSE~\cite{conneau2017word} that align the fastText monolingual embeddings pre-trained on Wikipedia corpora together in a single vector space. We collect 58 occupation words in Spanish from the occupation list provided by~\citet{font2019equalizing} and 23 in French (some occupations in French do not distinguish between two forms) for the bias mitigation experiments. We use WEAT word lists from~\citet{caliskan2017semantics} for male and female attribute words for our WEAT experiments and translate them to Spanish and French. For the word pair translation task, we use 7 common adjectives and pair every adjective with each occupation, resulting in 406 pairs for Spanish and 161 for French. 

\subsection{Monolingual Experiments}
\paragraph{Modified Word Embedding Association Test (MWEAT)} 
As discussed in Section \ref{sec:mweat}, we evaluate the gender bias using modified WEAT and we treat the absolute value of the sum of differences between mean cosine similarities of the two gender attributes for these occupation words $\left |\sum_{\vec{x}\in X}s(\vec{x},A,B)\right |$ as the association of male occupation target concepts $X$ with the gender attributes. Similarly, we use $\left |\sum_{\vec{y}\in Y}s(\vec{y},A,B)\right |$ as the association of female occupation words with the gender attributes. Finally our test statistic is 
\begin{equation*}
   \left | \left |\sum_{\vec{x}\in X}s(\vec{x},A,B)\right | - \left |\sum_{\vec{y}\in Y}s(\vec{y},A,B)\right | \right |,
\end{equation*}
which measures the difference in the association strength for two target sets with the two attribute sets occupation words. We then calculate the one-side p-value of the permutation test.

\begin{table*}[ht]
\centering
\begin{tabularx}{\linewidth}{c|c||c||c|c|c|c|c}
Task  & Metric               & Original                & Shift\_Ori              & Shift\_EN               & De-Align                & Hyrid\_Ori              & Hyrid\_EN               \\ \hline
\multicolumn{8}{c}{Word Translation} \\
\hline
EN$\rightarrow$ES & P@1/P@5     & 79.2/89.0          & \textbf{80.7/90.3}  & \textbf{80.7/90.3}  & 76.5/88.9          & \textbf{80.7/90.3}  & \textbf{80.7/90.3}  \\
ES$\rightarrow$EN & P@1/P@5     & 79.2/89.0          & 79.2/89.0           & 79.2/89.0           & \textbf{80.1/90.7} & 79.2/89.0           & 79.2/89.0           \\
EN$\rightarrow$FR &P@1/P@5     & 78.2/89.4          & \textbf{79.9/91.1} & \textbf{79.9/91.1}  & 74.3/87.8          & \textbf{79.9/91.1} & \textbf{79.9/91.1} \\
FR$\rightarrow$EN &P@1/P@5     & \textbf{76.1/88.1} & \textbf{76.1/88.1}  & \textbf{76.1/88.1} & 74.4/87.2          & \textbf{76.1/88.1}  & \textbf{76.1/88.1}  \\ \hline
\multicolumn{8}{c}{Word Pair Translation} \\ 
\hline
\multirow{4}{*}{EN$\rightarrow$ES} & ASD        & 0.1082                  & 0.0961                   & 0.0961                   & 0.0827                  & \textbf{0.0755}          & 0.0772                   \\
 & F\_MRR & 0.2073                  & 0.2507                   & 0.2507                   & 0.2919                  & 0.3450          & 0.3150                   \\
 & M\_MRR   & 0.6940                  & 0.6766                   & 0.6766                   & 0.6775                  & 0.6398          & 0.6696                   \\
 & MRR Diff   & 0.4867                  & 0.4259                   & 0.4259                   & 0.3856                  & \textbf{0.2949}          & 0.3546                   \\ \hline
\multirow{4}{*}{EN$\rightarrow$FR}  & ASD          & 0.1208                  & 0.1048                   & 0.1082                   & 0.0892                  & \textbf{0.0735}          & 0.0805                   \\
 & F\_MRR & 0.1663                  & 0.2101                   & 0.1943                   & 0.2679                  & 0.3128          & 0.2975                   \\
 & M\_MRR   & 0.6549                  & 0.6313                   & 0.6419                   & 0.6610                  & 0.6393         & 0.6467                   \\
 & MRR Diff   & 0.4886                  & 0.4212                   & 0.4476                   & 0.3931                  & \textbf{0.3265}          & 0.3492                   \\ 
\end{tabularx}
\caption{Results on word translation and word pair translation based on Spanish-English and French-English bilingual embeddings. 
We find that the original bilingual embeddings have a large discrepancy between the two genders, indicated by the average cosine similarity difference (ASD) and mean reciprocal ranks (MRR) difference. After applying the mitigation methods, both ASD and MRR difference drop. }%
\label{tab:bilingual_results}
\end{table*}


\paragraph{Word Similarity}
We test the quality of the embeddings after mitigation on the SemEval 2017 word similarity task~\cite{camacho2017semeval} for monolingual embeddings. This task evaluates how well the cosine similarity between two words correlates with a human-labeled score for which we report the Pearson correlation score.

\begin{figure}
	\includegraphics[width=\columnwidth]{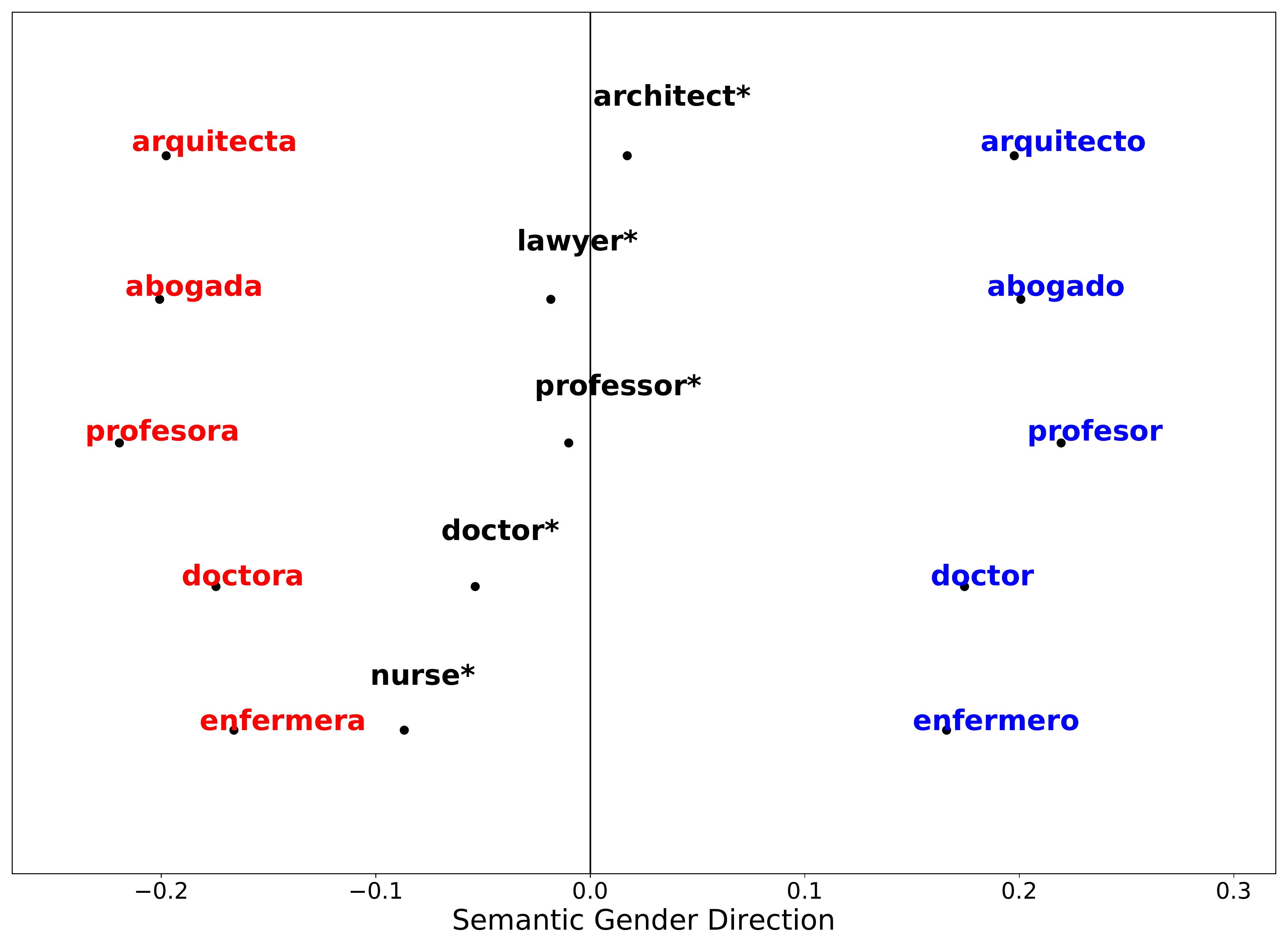}
	\centering
	\caption{Projections of occupations words on the semantic gender direction after hybrid mitigation with origin as the anchor point. The two forms of occupations are a lot more symmetric.}~\label{fig:viz_bilingual_after}
\end{figure}
\paragraph{Results}
Table~\ref{tab:mono_results} shows that Hybrid\_Ori significantly decreases the difference of association between two genders as indicated by MWEAT--Diff and p-value. Other methods only show marginal mitigation effects in terms of p-value.
We also find that the performance of our mitigation methods on word similarity is largely preserved as indicated by the Pearson correlation scores. 
\subsection{Cross-lingual Experiments} \label{bilingual_exp}
\paragraph{Word Translation}
We use word translation to examine whether our mitigation approaches preserve the utility of the bilingual word embediddings. This task aims to retrieve the translation of a source word in the target language. We use the test set provided by~\citet{conneau2017word}, which contains 200,000 randomly selected candidate words for 1,500 query words. We translate a query word and report the precision at $k$ (i.e., 
fraction of correct translations that are ranked not larger than $k$) by retrieving its $k$ nearest neighbours in the target language. 
We report both results with $k = 1$ and $k=5$.
\paragraph{Word Pair Translation by Analogy}
We propose a word pair translation task to evaluate the bias in gendered languages. In languages with grammatical gender, word forms of articles and adjectives need to agree with the gender of nouns they are associated with. For example, the word ``good'' has two forms in Spanish: ``bueno'' (masculine form) and ``buena'' (feminine form). Following the notion of analogy, we design an evaluation task to test how bilingual word embeddings translate an English occupation with the presence of another word.
Specifically, given an English word $E_i$ (a noun, an adjective or a verb), an English occupation word $E_o$, and the corresponding Spanish/French translation $S_i$ of $E_i$,
we adopt the analogy test ``$E_i$:$E_o$ = $S_i$:?'' (e.g., ``good:doctor = buena: ?'')  to rank all the words $S_o$ in Spanish/French based on the cosine similarity: $\cos(S_o, E_o - E_i + S_i)$. Ideally, the correct translation of $E_o$ that agree with gender of $S_i$ should be ranked on the top (``doctora'' in the above example). Quantitatively, we use the mean reciprocal ranks (MRR) to evaluate the performance of the ranking. If the bilingual embedding is not biased, the model should perform similarly on answering the translation of masculine and feminine occupations. Based on this, we use the gap between the performances on two form of genders to quantify the bias in the bilingual embedding space. Besides, we also report the average cosine similarity difference (ASD) of occupation words between two different gender forms.


\begin{figure}
	\includegraphics[width=\columnwidth]{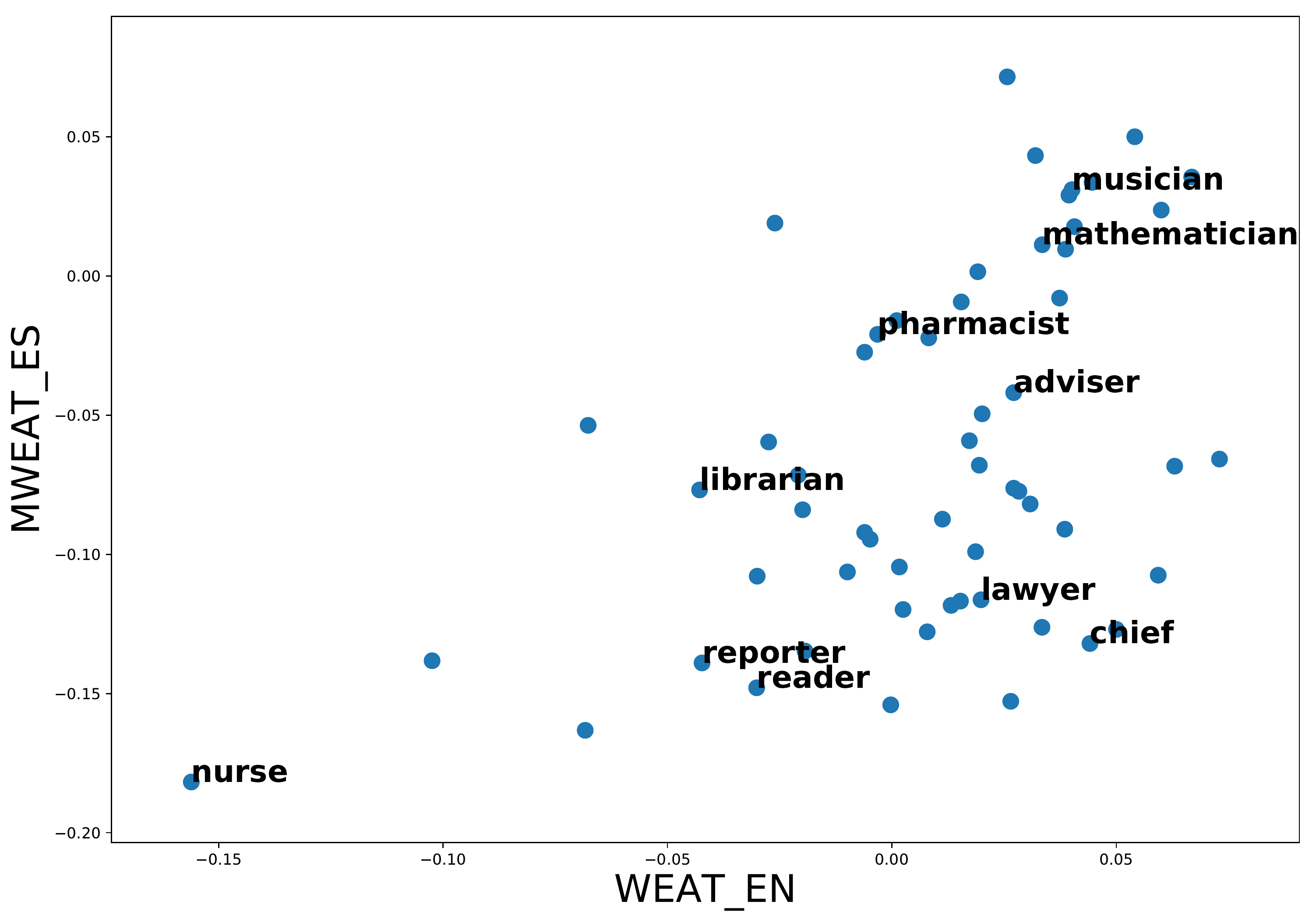}
	\vspace{-1em}
	\centering
	\caption{(M)WEAT scores for occupations in English and Spanish. For both axes, the smaller the number, the occupation is more leaning towards to female.}~\label{fig:ES_EN_corr}
	\vspace{-2em}
\end{figure}

\paragraph{Results}

Table~\ref{tab:bilingual_results} shows the results.  Results on word translation shows that the mitigation approaches do not harm the utility of bilingual word embedding and words in different languages still align well. 
From the results of word pair translation, the original bilingual embedding exhibits strong bias and the gap between two genders are around 0.49 in MRR in both ES-EN and FR-EN bilingual embeddings and the average cosine similarity difference is larger than 0.1. 
Hybrid\_Ori results in smallest difference for cosine similarity and MRR gap between two gender forms. However, the discrepancy between two forms is not completely removed.

To visualize the effect of bias mitigation, we project the bias-reduced embeddings of occupation words on the semantic gender direction as shown in Figure \ref{fig:viz_bilingual_before}.  Figure~\ref{fig:viz_bilingual_after} shows that after debiasing using Hybrid\_Ori, the two gender forms of occupation words are more symmteric with respect to the corresponding English embedding. 

\subsection{Bias Correlation between Languages}
Finally, we compute the WEAT/MWEAT scores\footnote{We use WEAT score for English and MWEAT for Spanish. To show the direction of bias, we 
take out the outer absolute function in Eq. \eqref{eq:mweat}.} for each occupation and show the commonality and difference of gender bias between English and Spanish embedding. Results in Figure~\ref{fig:ES_EN_corr} shows that 
occupations (``nurse'' and ``mathematician'') lean towards the same gender in both languages. However, some occupations (e.g., ``chef'') lean to different genders in different languages. In general, there is a strong correlation between gender bias in English and Spanish embedding (Spearman correlation=0.45 with p-value=0.0004).

\section{Related Work}\label{rel_work}
Previous work has proposed to quantify bias in English Embedding
definitions for gender bias in English word embeddings. 
Besides the aforementioned studies~\cite{kaiwei,caliskan2017semantics} in quantifying bias in English embedding,
\citet{katherine2017grammatical} examine grammatical gender bias in word embeddings of gendered languages by computing the WEAT association score~\cite{caliskan2017semantics} between gendered object nouns (e.g. moon-sun) and gender-definition words.
They propose to mitigate bias by applying lemmatization to remove gender information from the training corpus.
However, their focus is different from us as our approaches aim at keeping the grammatical gender information and only removing the bias in semantic genders. 
A few recent studies focus on measuring and reducing gender bias in contextualized word embeddings~\cite{ZWYCOC19,may2019measuring,basta2019evaluating}. 
However, they only focus on English embeddings in which the gender is mostly only expressed by pronouns~\cite{stahlberg2007representation}.
An interesting future direction is to extend their analyses to language with grammatical gender. 

Regarding bias mitigation appraoches, \citet{zhao2018learning} mitigate bias by saving one dimension of the word vector for gender.~\citet{bordia2019identifying} propose a regularization loss term for word-level language models.~\citet{zhang2018mitigating} use an adversarial network to mitigate bias in word embeddings. All these approaches consider only English embeddings. Moreover,~\citet{gonen2019lipstick} show that mitigation methods based on gender directions are not sufficient, since the embeddings of socially-biased words still cluster together.


Bias in word embedding may affect the downstream applications~\cite{ZWYOC18,rudinger2018gender,font2019equalizing}.
Besides the gender bias in word embeddings, implicit stereotypes have been shown in other real world applications, such as online reviews~\cite{wallace2016jerk}, advertisement~\cite{sweeney2013discrimination} and web search~\cite{kay2015unequal}. 

\section{Conclusion and Future Work}\label{conclusion}
We analyze gender bias in the embeddings of languages with grammatical gender and bilingual word embeddings that align such a language with a language without grammatical gender. 
We propose new methods to evaluate and mitigate gender bias for languages with grammatical gender and bilingual word embeddings and results show that our methods can mitigate gender bias while preserving the quality of original embeddings.

Directions for future work include testing monolingual and bilingual word embeddings on downstream tasks like machine translation to measure bias and also test the performance for mitigation methods. Moreover, the number of noun classes for other languages with grammatical gender ranges from two to several tens~\cite{corbett1991gender} and future work can extend methods proposed in this paper to address grammatical gender in languages with more gender forms.
\section{Acknowledgements}
This work was supported in part by DARPA (HR0011-18-9-0019). Ryan Cotterell was supported by a Facebook Fellowship. We acknowledge fruitful discussions with Fred Morstatter, Nanyun Peng, Ninareh Mehrab, Shunjie Wang, Yizhou Yao, and Aram Galstyan. We also thank anonymous reviewers for their comments.

\bibliography{emnlp-ijcnlp-2019}
\bibliographystyle{acl_natbib}



\end{document}